\documentclass{article}
\usepackage{ijcai19}

\usepackage{times}
\usepackage{soul}
\usepackage{url}
\usepackage[hidelinks]{hyperref}
\usepackage[utf8]{inputenc}
\usepackage[small]{caption}
\usepackage{graphicx}
\usepackage{amsmath}
\usepackage{amssymb}
\usepackage{booktabs}
\usepackage{algorithm}
\usepackage{algorithmic}
\usepackage{tabularx}
\title{Towards Diverse Paraphrase Generation Using Multi-Class Wasserstein GAN}

\author{
Zhecheng An$^1$
\and
Sicong Liu$^{2,3}$
\affiliations
$^1$Department of Electronic Engineering, Tsinghua University, Beijing, P. R. China.\\
$^2$Department of
Communication Engineering, Xiamen University,
Xiamen, P. R. China.\\
$^3$Key Laboratory of Digital Fujian on IoT
Communication, Architecture and Security Technology, Xiamen University,
Xiamen, P. R. China.
\emails
anzhecheng@gmail.com,
liusc@xmu.edu.cn
}

\begin{document}

\maketitle

\begin{abstract}
Paraphrase generation is an important and challenging natural language processing (NLP) task.
In this work, we propose a deep generative model to generate paraphrase with diversity.
Our model is based on an encoder-decoder architecture.
An additional transcoder is used to convert a sentence into its paraphrasing latent code.
The transcoder takes an explicit pattern embedding variable as condition,
so diverse paraphrase can be generated by sampling on the pattern embedding variable.
We use a Wasserstein GAN to align the distributions of the real and generated paraphrase samples.
We propose a multi-class extension to the Wasserstein GAN,
which allows our generative model to learn from both positive and negative samples.
The generated paraphrase distribution is forced to get closer to the positive real distribution,
and be pushed away from the negative distribution in Wasserstein distance.
We test our model in two datasets with both automatic metrics and human evaluation.
Results show that our model can generate fluent and reliable paraphrase samples
that outperform the state-of-art results, while also provides reasonable variability and diversity.
\end{abstract}

\section{Introduction}

Paraphrases are rewritten versions of text with different words or expressions while preserving the original
semantic.
The automatic paraphrase generation of a given sentence is an important NLP task, which can
be applied in many fields such as information retrieval, question answering, text summarization, dialogue system, etc.
A paraphrase generator is able to perform text reformulation on these systems to bring variation. Besides, the generated
paraphrases can be used as augmented data in many learning tasks such as text identification, classification and
inference. Therefore, the generation fidelity, naturalness and diversity play important roles in the
evaluation on a paraphrase generator system.

Paraphrase generation is a challenging task due to the complexity of human language. The recent progress of deep learning,
especially sequence-to-sequence (Seq2Seq) based models for text generation \cite{bahdanau2015neural,bowman2016generating},
have shown great advantages over the traditional rule-based \cite{mckeown1983paraphrasing} and statistic \cite{quirk2004monolingual} models.
Intuitively, a straightforward method to generate paraphrase is to train a Seq2Seq model to convert a sentence into its paraphrasing reference
using the maximum likelihood estimation (MLE), where the cross entropy loss is optimized \cite{prakash2016neural,cao2017joint}. This method is further extended in
\cite{ranzato2016sequence} to use metrics like
BLEU \cite{Papineni2002} and ROUGE \cite{lin2004rouge} as reward function of the
reinforcement learning algorithm. To mitigate the gap between lexical similarity and sematic similarity,
\cite{li2018paraphrase} replaces the lexical reward function with a trained evaluator and update it using inverse reinforcement learning.

The researches mentioned above focus on converting a sentence into a paraphrasing target optimized with various metrics.
However, there may be multiple possible paraphrases for one given sentence. Paraphrases of a certain sentence generated by humans
may differ from each other and contain wide linguistic variations, which cannot be captured by single-target transformation models.
Moreover, these models tend to generate sentences with high resemblance to the training samples, while other good semantically similar results
may be suppressed \cite{li2018paraphrase}. Therefore, in order to further exploit the variability and obtain diverse paraphrases,
it is necessary to generatively model the paraphrase distribution instead of a single target sample.

In order to model the distribution of the generated paraphrase, we introduce a random variable
as \emph{pattern embedding}. The generated results are explicit conditioning on the pattern
embedding variable. Therefore, the model can generate multiple results with diversity by sampling
on this variable.
To train such a generative model,
one existing work is the VAE-SVG \cite{Gupta2017} that uses a
conditional variation auto-encoder (VAE) \cite{kingma2013auto} for paraphrase generation.
However, in this paper, we exploit the adversarial generative network (GAN) \cite{goodfellow2014generative}
to model the paraphrase generation distribution as an alternative approach. 
Instead of using the KL-divergence to optimize the lower-bound in VAE,
the GAN uses adversarial training to directly align the generated distribution with
the real distribution, which is able to generate realistic results.

Applying GANs on text generation is non-trivial since text consists of discrete tokens
that are non-differentiable. We use the Gumbel-softmax \cite{Jang2016} as a continuous
approximation and use professor-forcing \cite{lamb2016professor} algorithm to match the hidden states of input
and paraphrasing sequences. In order to integrate professor-forcing in our model, we design an auto-encoder along with
a transcoder as the generator. The transcoder is a feedforward network that takes both the original sentence and the
pattern embedding as inputs, and outputs the paraphrase latent code.
A shared decoder is used to generate paraphrase sentence and decode the reference sample.

Specifically, we take advantage of the Wasserstein GAN (WGAN) \cite{Arjovsky2017} to train our paraphrase generation model
for better stability and convergence performance. We propose a multi-class extension to WGAN by using multiple critics to
measure the generated Wasserstein distance to different classes of samples. The multi-class WGAN enables our model
to learn paraphrase generation from both positive and negative samples.
The generated paraphrase distribution is forced to get closer to the
positive distribution and be pushed away from the negative distribution in Wasserstein distance,
which contributes to the generation
fluency and relevance.

Overall, the main contributions of this work are summarized as follows:
(1) We propose a generative model aiming at generating multiple paraphrases of a given sentence with diversity.
(2) With continuous approximation and professor-forcing, the model is trained with GAN to align the generated distribution
with the real distribution.
(3) We develop the multi-class WGAN that enables our model to learn from both positive and negative samples, which
promotes the generation fluency and relevance.

\section{Related Work}

\paragraph{Neural Paraphrase Generation:}
\cite{prakash2016neural} proposes a Seq2Seq paraphrase generation
model using residual stack LSTM and cross entropy loss.
\cite{cao2017joint} introduces an additional copying decoder for keywords extraction from the source.
\cite{xu2018d} uses a fixed vocabulary of rewrite patterns in the decoder to generate diverse paraphrases,
and the model is trained using MLE criterion by optimizing on selective patterns.
The evaluation of paraphrasing is studied in \cite{li2018paraphrase}, where a trained evaluator is used
as the reward function to train a paraphrase generation model with inverse reinforcement learning.
Besides the above transformation-based model, generative model to formulate the paraphrase generation
distribution is also proposed, such as the VAE-based VAE-SVG \cite{Gupta2017}.
In this paper, we use GAN as an alternative generative approach for paraphrase distribution modeling.
To the best of the authors' knowledge, this work is the first in literature that applies GAN in paraphrase generation.

\paragraph{Generative Adversarial Networks:} The main idea of
GAN \cite{goodfellow2014generative} is to train a generator and a discriminator that compete with each other,
forcing the generator to generate realistic outputs to fool the discriminator. In such way, the generated distribution
of GAN is forced to align with the real distribution. Various extensive algorithms to the vanilla GAN have
been proposed to handle different tasks. For example,
conditional GAN (CGAN) \cite{mirza2014conditional} is used to model conditional distribution by feeding
side information to the generator and discriminator.
In ACGAN \cite{odena2017conditional}, an auxiliary classifier is added to the discriminator to tackle the multi-class
generation problem.
WGAN \cite{Arjovsky2017} modifies the discriminator as the critic to measure Wasserstein distance instead of
Jensen-Shannon (JS) divergence, and achieves better training stability.

\paragraph{GAN-based Text Generation:} Since GANs achieve many success in image generation fields, several recent
researches focus on applying GAN in text generation. For example, \cite{hu2017toward} combines a discriminator with the VAE model
to generate text with controllable attribute. With non-parallel corpus, \cite{Shen2017} cross-align distributions
between two datasets with GAN to
perform style transfer. Such adversarial training technique is also used in unsupervised neural machine translation (NMT)
\cite{lample2017unsupervised} to match the encoded latent spaces of two languages.
For supervised NMT with pairwise samples, \cite{Wu2017} designs an Adversarial-NMT model using
GAN as a probabilistic transformer to process translation on parallel corpus.

\section{Method}

\subsection{Model Framework}

\begin{figure}
  \centering
  \includegraphics[width=0.45\textwidth]{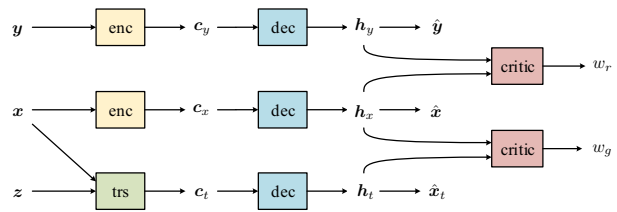}\\
  \caption{Overview of the proposed paraphrase generation framework.}\label{fig:framework}
\end{figure}

The overall framework of our proposed paraphrase generation model is shown in Figure \ref{fig:framework}.
The model consists of an auto-encoder, a transcoder and a critic. The auto-encoder is used to encode and reconstruct
the input and reference paraphrasing sentences.
The transcoder is a feedforward network that converts a sentence into its paraphrasing latent code, which is then decoded
with a decoder shared with the auto-encoder. Finally, the decoded paraphrase result is matched with the recovered real sample
using the critic.

\begin{figure*}[t]
  \centering
  \includegraphics[width=0.94\textwidth]{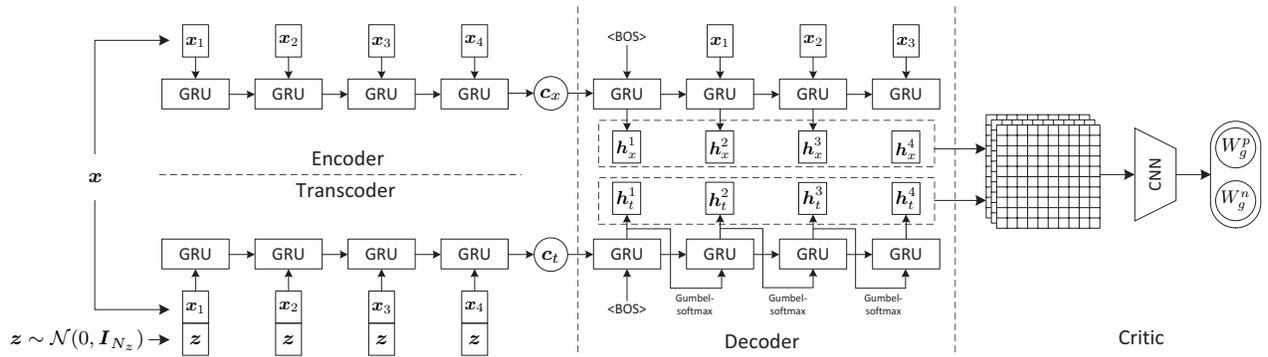}\\
  \caption{The structure of the proposed auto-encoder, transcoder and critic. }\label{fig:detail}
\end{figure*}

\paragraph{Auto-encoder:}
 Consider a pair of paraphrasing sentences $(\boldsymbol{x}, \boldsymbol{y})$, where $\boldsymbol{x}=\{x_1,x_2,\cdots,x_T\}$ and
$\boldsymbol{y}=\{y_1,y_2,\cdots,y_T\}$ are sequences of tokens. In the auto-encoder, $\boldsymbol{x}$ and $\boldsymbol{y}$ are encoded
into latent code $\boldsymbol{c}_x=\text{enc}_{\theta_e}(\boldsymbol{x})$ and $\boldsymbol{c}_y=\text{enc}_{\theta_e}(\boldsymbol{y})$,
where $\theta_e$ refers to the encoder parameter.
$\boldsymbol{c}_x$ and $\boldsymbol{c}_y$ are then decoded
by $\text{dec}_{\theta_d}(\cdot)$ with parameter $\theta_d$ to recover the original sequences
 as $\hat{\boldsymbol{x}}$ and $\hat{\boldsymbol{y}}$.
Gated recurrent unit (GRU) based recurrent neural networks (RNN) are used in the encoder and decoder. $\boldsymbol{h}_x$ and
$\boldsymbol{h}_y$ denote the decoding hidden states.
The auto-encoder is trained in a teacher-forcing pattern, where ground truth samples are fed into the decoder
every time step during training.
The training objective of auto-encoder is to minimize the reconstruction loss,
 which is the sum of token-level cross-entropy loss in this paper, i.e.
\begin{align}
\mathcal{L}_{\text{AE}}(\theta_e, \theta_d)=\mathbb{E}_{(\boldsymbol{x}, \boldsymbol{y})}
\big[&-\log p_{\text{dec}}(\boldsymbol{x}|\text{enc}_{\theta_e}(\boldsymbol{x}); \theta_d) \nonumber\\
&-\log p_{\text{dec}}(\boldsymbol{y}|\text{enc}_{\theta_e}(\boldsymbol{y}); \theta_d)\big]. \label{eqn:ae_loss}
\end{align}

\paragraph{Transcoder:} The main purpose of this work is to model the paraphrase distribution of a given sentence
$p(\boldsymbol{x})$ instead of a transformation function $f(\boldsymbol{x})$. In order to achieve this goal,
we introduce a random variable $\boldsymbol{z}$ as the pattern embedding variable, and perform paraphrase generation
conditioning on $\boldsymbol{z}$. Therefore, the paraphrase distribution can be derived as $p(\boldsymbol{x})=f(\boldsymbol{x}|\boldsymbol{z})$,
and diverse paraphrase of $\boldsymbol{x}$ can be generated by sampling on $\boldsymbol{z}$.
Specifically, we use a feedforward GRU network as the transcoder as shown in Figure \ref{fig:detail},
which takes $\boldsymbol{x}$ and $\boldsymbol{z}$ as inputs, and
convert $\boldsymbol{x}$ into its paraphrasing form in the latent space, i.e.
$\boldsymbol{c}_t = \text{trs}_{\theta_t}(\boldsymbol{x}|\boldsymbol{z})$.
In particular, $\boldsymbol{z}$ is a $N_z$-dimensional random vector sampled from a standard normal distribution $\mathcal{N}(0, \boldsymbol{I}_{N_z})$,
and concatenated with each token in $\boldsymbol{x}$ for the transcoder input.
The latent code
$\boldsymbol{c}_t$ is decoded with a decoder that shares weight with the auto-encoder to output
the final paraphrase sequence $\hat{\boldsymbol{x}}_t$, where $\boldsymbol{h}_t$ refers to the corresponding decoding hidden
states. As shown in Figure \ref{fig:detail}, $\boldsymbol{c}_t$ is decoded in a free-run mode,
where the output of last state is used as the input
of the next state.

\paragraph{Critic:}
In order to train the output distribution of the paraphrase generation, we apply WGAN in our model, where
a critic is implemented aiming at distinguishing generated fake samples from real samples.
With a decoded sentence $\hat{\boldsymbol{x}}$ as condition, the critic is trained to detect whether a sentence is the real
paraphrase of $\hat{\boldsymbol{x}}$. The critic outputs for real and generated samples are denoted as
$w_r=f_{\theta_c}(\hat{\boldsymbol{y}}|\hat{\boldsymbol{x}})$
and $w_g=f_{\theta_c}(\hat{\boldsymbol{x}}_t|\hat{\boldsymbol{x}})$, with parameter $\theta_c$. The structure
of the critic is detailed in Section \ref{sec:critic}.

\subsection{Multi-class Wasserstein GAN}

WGAN \cite{Arjovsky2017}, as an improved GAN algorithm, 
utilizes the Wasserstein distance
instead of JS-divergence to achieve better stability and avoid mode collapse problems. 
Given the distribution of real and generated samples as $\mathbb{P}_r$ and $\mathbb{P}_{\theta_g}$,
the Wasserstein distance between the two distributions is defined as
\begin{equation}
W(\mathbb{P}_r, \mathbb{P}_{\theta_g}) = \max_{\theta_c \in \mathcal{W}}
\mathbb{E}_{x\sim \mathbb{P}_r} \big[f_{\theta_c}(x)\big]
 - \mathbb{E}_{x\sim \mathbb{P}_{\theta_g}} \big[f_{\theta_c}(x)\big],
\end{equation}
where $\{f_{\theta_c}\}_{\theta_c\in \mathcal{W}}$ is the family of all $K$-Lipschitz functions $f:\mathcal{X}\mapsto \mathbb{R}$.
The critic maps distributions into Wasserstrin distance, which acts differently as the discriminator in vanilla GAN.
Thus, auxiliary classifier can not be directly integrated into the critic as the AC-GAN \cite{odena2017conditional}
to handle multi-class generation problem. Therefore, we propose an alternative approach as follows.

We consider $N$ classes in the real samples, each with distribution $\mathbb{P}_r^{(i)}, 1 \leqslant i \leqslant N$.
For a certain generator with distribution $\mathbb{P}_{\theta_g}$, we use a critic with $N$ outputs to
meassure the Wasserstein distance between $\mathbb{P}_{\theta_g}$ and $\mathbb{P}_r^{(i)}$ respectively, i.e.
\begin{equation}\label{eqn:wass_distance}
W(\mathbb{P}_r^{(i)}, \mathbb{P}_{\theta_g}) = \max_{\theta_c \in \mathcal{W}}
\mathbb{E}_{x\sim \mathbb{P}_r^{(i)}} \big[f^{(i)}_{\theta_c}(x)\big]
 - \mathbb{E}_{x\sim \mathbb{P}_{\theta_g}} \big[f^{(i)}_{\theta_c}(x)\big].
\end{equation}
Suppose we are training a generator to generate samples of class $i$.
The generated distribution should have minimized
Wasserstein distance to $\mathbb{P}_r^{(i)}$,
while its Wasserstein distance to another class $\mathbb{P}_r^{(j)}\ (j\neq i)$ should exceed a margin
in order to be distinguishable across classes.
Therefore, we redefine the generator loss as
\begin{multline}\label{eqn:g_loss}
\mathcal{L}_g^{(i)}(\theta_g^{(i)})=(1-\beta)W(\mathbb{P}_r^{(i)}, \mathbb{P}_g^{(i)}) + \\
\frac{\beta}{N-1}\sum_{j\neq i}\Big[W(\mathbb{P}_r^{(i)}, \mathbb{P}_{\theta_g}^{(i)})
-W(\mathbb{P}_r^{(j)}, \mathbb{P}_{\theta_g}^{(i)})+\alpha\Big]_+,
\end{multline}
where $[\cdot]_+$ stands for a ReLU.
Eqn. (\ref{eqn:g_loss}) contains a term motivated by the triplet loss \cite{Schroff2015}, where we use
the Wasserstein distance between distributions to replace the L2 distance between samples. $\alpha$ refers to
the enforced margin, and $\beta$ refers to the weight on the negative loss.

In the paraphrase generation problem, some datasets contain both positive and negative samples. With
multi-class WGAN, the generator is able to learn from the positive samples to generate flexible paraphrases,
while also exploits from negative sample to improve the generation reliability. Given the real
positive and negative distributions as $\mathbb{P}_r^{(p)}$ and $\mathbb{P}_r^{(n)}$, the paraphrase
generator loss is formulated as
\begin{multline}
\mathcal{L}_g(\theta_t, \theta_d)=(1-\beta)W(\mathbb{P}_{r}^{(p)}, \mathbb{P}_{\theta_t, \theta_d})+ \\
\beta\big[W(\mathbb{P}_{r}^{(p)}, \mathbb{P}_{\theta_t, \theta_d}) -
W(\mathbb{P}_{r}^{(n)}, \mathbb{P}_{\theta_t, \theta_d}) + \alpha \big]_+ . \label{eqn:g_loss_final}
\end{multline}

\subsection{Continuous Approximation}

Applying adversarial training algorithm on RNN based text generator is hard since the generated sequence
is discrete and non-differentiable. One approach to tackle this problem is to use REINFORCE \cite{Sutton2000} algorithm.
However, the sampling-based gradient estimation suffers from high variance and unstable
training. Instead, we use the Gumbel-softmax \cite{Jang2016} trick as a continuous approximation
to handle the discrete sequence generation problem. In the decoding process of the paraphrase sequence generation, we use the
Gumbel-softmax distribution to replace the sampled token feeding to the next RNN step, i.e.
\begin{equation}\label{eqn:gumbel}
y_i = \frac{\exp[(\log\pi_i + g_i) / \tau]}{\sum_{j=1}^V\exp[(\log\pi_j + g_j) / \tau]}, \quad\text{for}\ 1\leqslant i \leqslant V
\end{equation}
where $[\pi_1,\cdots,\pi_V]$ is the probabilities of decoding tokens, $V$ is the vocabulary size,
$\tau > 0$ is a temperature parameter, and $g_i\sim \text{Gumbel}(0, 1)$ distribution.
Such reparameterization trick
provides a reasonable approximation and
makes the generator differentiable that allows gradients to back-propagate in training process.

\subsection{Critic Model}
\label{sec:critic}

Motivated by \cite{Shen2017}, we use professor-forcing \cite{lamb2016professor} to match the decoding hidden states
of auto-encoder and paraphrase generator, since they share the same
decoder parameters.
The hidden states of the paraphrase generator are trained to be
indistinguishable from hidden states of the teacher-forced auto-encoder.
By using hidden states as critic input, the Wasserstein distance defined in
Eqn. (\ref{eqn:wass_distance}) is reformulated as
\begin{equation}\label{eqn:wass_dist_cal}
\resizebox{1.\linewidth}{!}{$
W_{\theta_c}(\mathbb{P}_r^{(i)}, \mathbb{P}_{\theta_t,\theta_d}) =
\hspace{-0.5em}\mathop{\mathbb{E}}\limits_{\boldsymbol{h}_y \sim \mathbb{P}_r^{(i)}}
\big[f_{\theta_c}^{(i)}(\boldsymbol{h}_y | \boldsymbol{h}_x)\big] -
\hspace{-0.5em}\mathop{\mathbb{E}}\limits_{\boldsymbol{h}_t \sim \mathbb{P}_{\theta_t, \theta_g}}\hspace{-0.5em}
\big[f_{\theta_c}^{(i)}(\boldsymbol{h}_t | \boldsymbol{h}_x)\big]
$}
\end{equation}
where $\theta_c=\arg\min\mathcal{L}^{(i)}_c(\theta_c)$.
In order to enforce the Lipschitz constraint to the WGAN critic, we use the recently proposed
gradient penalty method \cite{Gulrajani2017}. A penalty term on gradient norm is added to the
critic loss, i.e.
\begin{multline}
\hspace{-0.5em}
\mathcal{L}^{(i)}_c(\theta_c) =
\hspace{-0.5em}\mathop{\mathbb{E}}\limits_{\boldsymbol{h}_t \sim \mathbb{P}_{\theta_t, \theta_g}}
\big[f_{\theta_c}^{(i)}(\boldsymbol{h}_t | \boldsymbol{h}_x)\big]
 - \hspace{-0.5em}\mathop{\mathbb{E}}\limits_{\boldsymbol{h}_y \sim \mathbb{P}_r^{(i)}}
\big[f_{\theta_c}^{(i)}(\boldsymbol{h}_y | \boldsymbol{h}_x)\big]
 \\ + \lambda \mathop{\mathbb{E}}\limits_{\hat{\boldsymbol{h}}\sim\mathbb{P}_{\hat{\boldsymbol{h}}}}
 \big[(\|\nabla_{\hat{\boldsymbol{h}}} f_{\theta_c}^{(i)}
 (\hat{\boldsymbol{h}}|\boldsymbol{h}_x)\|_2-1)^2 \big]\label{eqn:critic_loss}
\end{multline}
for $i=p$ or $n$, and $\hat{\boldsymbol{h}}$ is sampled randomly from linear interpolation
of real and generated samples.

We use a CNN model for the critic. For hidden states $\boldsymbol{h}_x$ and $\boldsymbol{h}$
($\boldsymbol{h}=\boldsymbol{h}_y$ or $\boldsymbol{h}_t$), we combine the two tensors into a 2-dimensional image like
representation.
For the $i$-th hidden state $\boldsymbol{h}_{x,i}$ in $\boldsymbol{h}_x$
and $j$-th hidden state $\boldsymbol{h}_{j}$ in $\boldsymbol{h}$,
the two hidden state vectors and their element-wise difference and product
are concatenated together forming a feature map as
\begin{equation}\label{eqn:feat_map}
\boldsymbol{w}_{i,j} = [\boldsymbol{h}_{x,i}^T, \boldsymbol{h}_{j}^T, |\boldsymbol{h}_{x,i} - \boldsymbol{h}_{j}|^T,
(\boldsymbol{h}_{x,i} \odot \boldsymbol{h}_{j})^T]^T
\end{equation}
The feature map is then fed into a CNN feature extraction network proposed in \cite{Gong2017},
which consists of several DenseNet \cite{Huang2016} blocks and transition blocks. The extracted features
are followed by an MLP to output the final estimation of Wasserstein distances.

\begin{algorithm}[tb]
\footnotesize
\caption{Proposed paraphrase generation algorithm}
\label{alg:algorithm}
\textbf{Input}: Positive and negative paraphrase sentences pair distributions $\mathbb{P}_r^{(p)}$
and $\mathbb{P}_r^{(n)}$, parameters $\alpha$, $\beta$, $\lambda$ and $\tau$
\begin{algorithmic}[1] 
\STATE Initialize $\theta_e$, $\theta_d$, $\theta_t$, $\theta_c$
\REPEAT{}
\STATE Sample mini-batches of $(\boldsymbol{x}^{(p)}, \boldsymbol{y}^{(p)})$
and $(\boldsymbol{x}^{(n)}, \boldsymbol{y}^{(n)})$ from $\mathbb{P}_r^{(p)}$ and $\mathbb{P}_r^{(n)}$, respectively
\STATE Sample a random pattern embedding $\boldsymbol{z}\sim\mathcal{N}(0,\boldsymbol{I}_{N_z})$
\STATE Compute paraphrase latent code $\boldsymbol{c}_t = \text{trs}_{\theta_t}(\boldsymbol{x}^{(p)}|\boldsymbol{z})$
\STATE Compute the paraphrase decoding hidden states in free-run mode with Gumbol-softmax approximation
$\boldsymbol{h}_t=\text{dec}_{\theta_d}(\boldsymbol{c}_t)$
\FOR{$i=p,n$}
\STATE Compute the encoder latent code $\boldsymbol{c}^{(i)}_{x} =\text{enc}_{\theta_e}(\boldsymbol{x}^{(i)})$ and
$\boldsymbol{c}^{(i)}_{y} =\text{enc}_{\theta_e}(\boldsymbol{y}^{(i)})$
\STATE Compute decoder hidden states in teacher-forcing mode as
$\boldsymbol{h}_x^{(i)}=\text{dec}_{\theta_d}(\boldsymbol{c}_x^{(i)})$ and
$\boldsymbol{h}_y^{(i)}=\text{dec}_{\theta_d}(\boldsymbol{c}_y^{(i)})$
\STATE Compute the auto-encoder loss $\mathcal{L}^{(i)}_{\text{AE}}(\theta_e,\theta_d)$ by Eqn. (\ref{eqn:ae_loss})
\STATE Combine hidden states as Eqn. (\ref{eqn:feat_map}), and compute critic output
$f_{\theta_c}^{(i)}(\boldsymbol{h}_t | \boldsymbol{h}_x^{(i)})$ and
$f_{\theta_c}^{(i)}(\boldsymbol{h}_y^{(i)} | \boldsymbol{h}_x^{(i)})$
\STATE Compute Wasserstein distance $W(\mathbb{P}_r^{(i)}, \mathbb{P}_{\theta_t,\theta_d})$
by Eqn. (\ref{eqn:wass_dist_cal}) and
critic loss $\mathcal{L}^{(i)}_{c}(\theta_c)$ by Eqn. (\ref{eqn:critic_loss})
\ENDFOR
\STATE Compute the paraphrase generator loss $\mathcal{L}_g(\theta_t, \theta_d)$ by Eqn. (\ref{eqn:g_loss_final})
\STATE Update $\{\theta_e, \theta_d, \theta_t\}$ by gradient descent on loss
\begin{equation*}
\mathcal{L}_{\text{g}} + \frac{1}{2}(\mathcal{L}_{\text{AE}}^{(n)}+\mathcal{L}_{\text{AE}}^{(p)})
\end{equation*}
\STATE Update $\theta_c$ by gradient descent on loss $\mathcal{L}_{\text{c}}=\mathcal{L}^{(p)}_{\text{c}}+\mathcal{L}^{(n)}_{\text{c}}$
\UNTIL{convergence}
\STATE \textbf{return} paraphrase generation model $\text{dec}_{\theta_d}[\text{trs}_{\theta_g}(\boldsymbol{x}|\boldsymbol{z})]$
\end{algorithmic}
\end{algorithm}

The overall training procedure of the proposed paraphrase generation model is detailed in Algorithm \ref{alg:algorithm}.

\section{Experiments}

\subsection{Datasets}

We train and evaluate our paraphrase generation model on the \textbf{Quora}
question pairs \footnote{https://www.kaggle.com/c/quora-question-pairs} dataset
and the \textbf{MSCOCO} \footnote{http://cocodataset.org} dataset. The Quora dataset
contains question pairs with human annotations originally aiming for paraphrase identification.
Therefore, besides the positive paraphrase examples, Quora dataset also contains non-trivial
negative examples, in which a pair of questions may share similar words but have different meanings. These
negative examples are helpful for our proposed multi-class WGAN model.
The Quora dataset consists of over 400K question pairs, after filtering question over 20 words,
we get 126K positive samples and 184K negative samples for the training set. For testing and
validation, we use two sets each with 4.5K positive samples. 
The MSCOCO contains an image captioning dataset with about 120K images with each having 5 human annotated captions,
which are used by some previous works \cite{Gupta2017} as a paraphrase dataset .
In this paper, we sample 75K pairs of captions to identical images in MSCOCO as the positive training set. We also randomly sample
another 75K pairs of captions of different images as negative set. Each of the testing and validation sets we use consists
of 20K samples of caption pairs. Since MSCOCO dataset dose not contain annotated negative samples, we only use it
to demonstrate the fidelity of our model across datasets.
The statistics of the two datasets used in this paper is presented in Table \ref{tab:datasets}.
\begin{table}
\centering
\scalebox{0.8}{
\begin{tabular}{lccccc}
\toprule
& \multicolumn{3}{c}{Generator} & \multicolumn{2}{c}{Critic}\\
\cmidrule(lr){2-4}\cmidrule(lr){5-6}
Dataset          & \#Train   & \#Test & \#Validation & \#Positive & \#Negative \\
\midrule
Quora    & 126K  & 4.5K   & 4.5K   & 126K & 184K   \\
MSCOCO       & 75K  & 20K  & 20K  & 75K & 75K   \\
\bottomrule
\end{tabular}
}
\caption{Statistics of datasets.}
\label{tab:datasets}
\end{table}

\subsection{Training Details}


We use the 300-dimensional pre-trained GloVe \footnote{https://nlp.stanford.edu/projects/glove/} word embeddings in our model.
The max length of input and output sentence is set as 20.
We implement the encoder, transcoder and decoder using RNNs with GRU cells.
The encoder and transcoder are two 2-layers bidirectional GRU networks with inner-attention, and the decoder is
a single-layer GRU network. The sizes of all the GRU hidden states are 512. The dimension of pattern embedding
is 128.
The DenseNet blocks and transition blocks in the critic are implemented the same as \cite{Gong2017},
except all the activation units are replaced by Leakly-ReLU. 

Before the adversarial training, we firstly pre-train the auto-encoder and the transcoder with the
MLE metric. The auto-encoder RNN is pre-trained in the teacher-forcing mode.
However, the transcoder needs to be trained in free-run mode, where
the Gumbel-softmax distribution of last state output given by Eqn. (\ref{eqn:gumbel}) is used as the next step input.


\section{Results and Analysis}

\begin{table}
\centering
\scalebox{0.8}{
\begin{tabular}{lcccc}
\toprule
& \multicolumn{4}{c}{Quora} \\
\cmidrule(lr){2-5}
Models           & BLEU   & ROUGE-1 & ROUGE-2 & METEOR \\
\midrule
Residual LSTM    & 29.63  & 58.89   & 30.72   & 31.62   \\
VAE-SVG          & 26.58  & 50.92   & 23.44   & 26.36   \\
Adversarial NMT  & 30.57  & 55.95   & 31.00   & \textbf{33.56}   \\
\midrule
MC-WGAN (average)& 27.54  & 56.45   & 27.75   & 28.14   \\
MC-WGAN (best)   & \textbf{32.33}  & \textbf{62.66}   & \textbf{36.06}   & 33.16   \\
\bottomrule
\end{tabular}
}
\caption{Automatic results on the Quora dataset.}
\label{tab:quora_result}
\end{table}
\begin{table}[t]
\centering
\scalebox{0.8}{
\begin{tabular}{lcccc}
\toprule
& \multicolumn{4}{c}{MSCOCO} \\
\cmidrule(lr){2-5}
Models           & BLEU   & ROUGE-1 & ROUGE-2 & METEOR \\
\midrule
Residual LSTM    & 21.90  & 33.21   & 11.53   & 16.27   \\
VAE-SVG          & 21.92  & 36.32   & 10.72   & 16.05   \\
Adversarial NMT  & 21.68  & 36.01   & 11.75   & 17.16   \\
\midrule
MC-WGAN (average)& 22.22  & 35.31   & 11.52   & 15.63   \\
MC-WGAN (best)   & \textbf{27.83}  & \textbf{48.42}   & \textbf{22.93}   & \textbf{22.78}   \\
\bottomrule
\end{tabular}
}
\caption{Automatic results on the MSCOCO dataset.}
\label{tab:mscoco_result}
\end{table}

\subsection{Baselines}

We compare the results of our proposed model with several existing paraphrase generation models, i.e.
residual LSTM \cite{prakash2016neural} (with two layers), VAE-SVG \cite{Gupta2017} and Adversarial NMT \cite{Wu2017}.
The reinforcement learning based Adversarial-NMT model is originally used in machine translation. We use it as a paraphrase generation model
by sharing vocabulary and word embeddings between the source and target languages. 
These models represent the typical approaches of neural paraphrase generation, and we use them as baselines to evaluate
our proposed model.

\subsection{Automatic Evaluation}

\begin{table}[t]
\centering
\scalebox{0.8}{
\begin{tabular}{lcc}
\toprule
Models           & Relevance   & Fluency  \\
\midrule
VAE-SVG          & 3.75        & 4.07       \\
\midrule
MC-WGAN          & \textbf{4.09}        & \textbf{4.22}       \\
\midrule\midrule
Reference & 4.88 &4.95  \\
\bottomrule
\end{tabular}
}
\caption{Human evaluation results on Quora dataset.}
\label{tab:human}
\end{table}

\begin{table*}[t]
\renewcommand\arraystretch{1.19}
\centering
\scalebox{0.8}{
\begin{tabular}{|p{11em}|p{11em}|p{11em}|p{11em}|p{11em}|}
\hline
\hspace{4em} Input & \hspace{3em} Reference & \hspace{2.5em} Generated $\boldsymbol{z}_1$ &
\hspace{2.5em} Generated $\boldsymbol{z}_2$ & \hspace{2.5em} Generated $\boldsymbol{z}_3$ \\
\hline
how do you start making money? &
what should i do to earn some more money? &
how do i make money through youtube? &
how do i make money from home? &
what are some ways to make money online? \\
\hline
how effective is scrapping 500 and 1000 rupee notes ? will it reduce black money? &
how will the ban on 500 and 1000 rupee note stop black money? &
how will banning 500 and 1000 rupee notes affect black money? &
how will the demonetization of 500 and 1000 rupee notes help indian economy? &
how will the ban of 500 and 1000 rupee notes help indian economy? \\
\hline
what are the worst mistakes of your life? &
what is the worst thing you did by mistake in your life? &
what is the worst mistake you have in your life? &
what was the most embarrassing moment of your life? &
what has been the worst experience of your life? \\
\hline
\end{tabular}
}
\caption{Some examples of paraphrases generated on Quora dataset.}
\label{tab:quora_demo}
\end{table*}
\begin{table*}[t]
\renewcommand\arraystretch{1.19}
\centering
\scalebox{0.8}{
\begin{tabular}{|p{11em}|p{11em}|p{11em}|p{11em}|p{11em}|}
\hline
\hspace{4em} Input & \hspace{3em} Reference & \hspace{2.5em} Generated $\boldsymbol{z}_1$ &
\hspace{2.5em} Generated $\boldsymbol{z}_2$ & \hspace{2.5em} Generated $\boldsymbol{z}_3$ \\
\hline
a group of kids playing a game of baseball. &
the young boys are playing a game of baseball in the park. &
a group of young children playing a game of baseball &
a group of baseball players playing a game on the playground. &
three young children playing baseball on a baseball team. \\
\hline
a man playing tennis going for a low ball &
a tennis player with a racket hitting the ball &
a man in a tennis court about to hit a tennis ball. &
a tennis player in a defensive stance to hit a ball with a racket. &
a man in a tennis court gets ready to hit a ball. \\
\hline
small pieces of cake have been arranged on a plate &
chocolate dessert bars covered in frosting and sprinkles. &
three pieces of cake are on a plate with a cut of syrup. &
two pieces of cake are on a plate with strawberries. &
three cakes on a plate that have been sliced on top. \\
\hline
\end{tabular}
}
\caption{Some examples of paraphrases generated on MSCOCO dataset.}
\label{tab:mscoco_demo}
\end{table*}

We first conduct automatic quantitative evaluations to compare the paraphrase
generation performance using BLEU-4 \cite{Papineni2002}, ROUGE-1 and ROUGE-2 \cite{lin2004rouge},
and METEOR \cite{Denkowski2015}. These metrics mainly consider the precision and recall of n-grams
between the generated sentences and the references. Synonyms from WordNet are also considered in METEOR.
However, these lexical metrics are not ideally suitable for the evaluation of paraphrase generation, because good paraphrasing
examples may exist besides the given references. This occurs more seriously when a model is aiming at generating diverse
paraphrasing samples, since the generation diversity is traded-off to the accuracy on specific references.
Therefore, we only use these automatic metrics as part of our evaluation along with human evaluation.

Table \ref{tab:quora_result} and \ref{tab:mscoco_result} show the performance of models on the Quora and MSCOCO datasets respectively.
Since our proposed multi-class WGAN (MC-WGAN) model can generate multiple paraphrases of a given sentence,
we list the average and best results separately. Table \ref{tab:quora_result} shows that the best performance of our model
outperforms the baseline models in all the considered metrics except for METEOR, which is close to the Adversarial NMT.
This indicates our model has the ability to generate result close to the reference,
i.e. the best results with respect to the ground truth are within our generation distribution. This is also shown by Table \ref{tab:mscoco_result}
on the MSCOCO dataset. Table \ref{tab:quora_result} shows the average performance of our model is no better than the Residual LSTM and
Adversarial NMT on Quora dataset, because both Residual LSTM and Adversarial NMT model contain MLE terms in their generator loss
and tend to generate samples close to the ground truth. However, with the help of GAN,  our model mainly focuses on a distribution perspective.
VAE-SVG model is also enabled to generate multiple paraphrases.
Table \ref{tab:quora_result} shows the average performance of our model outperforms VAE-SVG on Quora
dataset, since the MC-WGAN learns from both positive and negative samples. However, on the MSCOCO dataset, performance gains
only show on BLEU and ROUGE-2, because the negative samples are randomly selected.

\subsection{Human Evaluation}

Table \ref{tab:human} shows the human evaluation performance on Quora dataset, where we mainly compare our model
against the VAE-SVG model since both the two are generative models that generate diverse paraphrase results.
We randomly choose 200 sentences generated by each model, and assign all the tasks to 3 individual human evaluators
to score ranging from 1 to 5 according to the relevance and fluency of each paraphrasing pair. (1 refers to the worst and 5 refers to the best).
Results show that our proposed model generates better paraphrasing samples than the VAE-SVG model in both
relevance and fluency metrics on Quora dataset. This is partially because our model succeeds in taking advantage of the negative samples to learn
better generation distribution.

\subsection{Generation Diversity}


Table \ref{tab:quora_demo} and \ref{tab:mscoco_demo} show some examples of paraphrases generated
with our model on Quora and MSCOCO dataset. By sampling on the pattern embedding vector $\boldsymbol{z}$,
the model is able to generate multiple paraphrases of a given sentence. The shown examples capture the accurate
semantic of the input sentences, while provide reasonable variation in the paraphrasing outputs. The results on MSCOCO
show greater variation in the paraphrases than the Quora dataset. This is because different captions may describe one
image from different aspects, which means the captions may not be strictly semantically identical as the human
annotated samples in Quora dataset. Our model is able to capture this feature in the generation phase, which
leads the generator to add more variational details in the results. 

\section{Conclusions}

In this paper, we have proposed an alternative deep generative model based on WGAN to
generate paraphrase of given text with diversity.
We build our model with an auto-encoder along with a transcoder.
The transcoder is conditioning on an explicit pattern embedding variable, and transcodes
an input sentence into its paraphrasing term in latent space.
Consequently, diverse paraphrases can be generated by sampling on the pattern embedding variable.
We apply WGAN to force the decoding paraphrase distribution to match the real distribution.
By extending WGAN to multiple class generation,
the generative model is enabled to learn from both the positive and negative real distributions for better
generation quality.
The proposed model is evaluated on two datasets with both automatic metrics and human evaluation.
Results show that our proposed model can generate fluent and reliable paraphrase samples
that outperform the state-of-art results, while also provides reasonable variability and diversity at the same time.
Our model provides a new baseline in generative paraphrase modeling.
The proposed model with the multi-class WGAN
algorithm can be potentially applied in may other text generation tasks with multiple labels, such as natural language inference
generation, in the future works.

\bibliographystyle{named}
\bibliography{nlg}

\end{document}